\documentclass{bmvc2k}
\usepackage{graphicx}
\usepackage{tikz}
\usepackage{comment}
\usepackage{amsmath,amssymb} 
\usepackage{color}

\usepackage{bm}
\usepackage{multirow}
\usepackage{xcolor}

\usepackage{booktabs}

\title{Inharmonious Region Localization via Recurrent Self-Reasoning
}

\addauthor{Penghao Wu}{wupenghaocraig@sjtu.edu.cn}{1}
\addauthor{Li Niu$^\ast$}{ustcnewly@sjtu.edu.cn}{1}
\addauthor{Jing Liang}{leungjing@sjtu.edu.cn}{1}
\addauthor{Liqing Zhang}{zhang-lq@cs.sjtu.edu.cn}{1}

\addinstitution{
 MoE Key Lab of Artificial Intelligence \\
 Shanghai Jiao Tong University \\
 Shanghai, China
}

\runninghead{Wu ET AL.}{Inharmonious Region Localization via RSR}

\begin{document}

\maketitle

\begin{abstract}
Synthetic images created by image editing operations are prevalent, but the color or illumination inconsistency between the manipulated region and background may make it unrealistic. Thus, it is important yet challenging to localize the inharmonious region to improve the quality of synthetic image. Inspired by the classic clustering algorithm, we aim to group pixels into two clusters: inharmonious cluster and background cluster by inserting a novel Recurrent Self-Reasoning (RSR) module into the bottleneck of UNet structure. The mask output from RSR module is provided for the decoder as attention guidance. Finally, we adaptively combine the masks from RSR and the decoder to form our final mask. Experimental results on the image harmonization dataset demonstrate that our method achieves competitive performance both quantitatively and qualitatively. 
\end{abstract}

\section{Introduction} \label{sec:intro}
\noindent Thanks to the rapid development of digital photography and editing software, synthetic images created by image editing operations (\emph{e.g.}, crop and paste, appearance adjustment) are prevalent in our daily lives. However, one crucial issue of some synthetic images is that the color and illumination characteristics of the manipulated regions are inconsistent with other regions, which could severely degrade the quality of synthetic images.

\par
\begin{figure}[t]
    \centering
    \includegraphics[scale = 0.22]{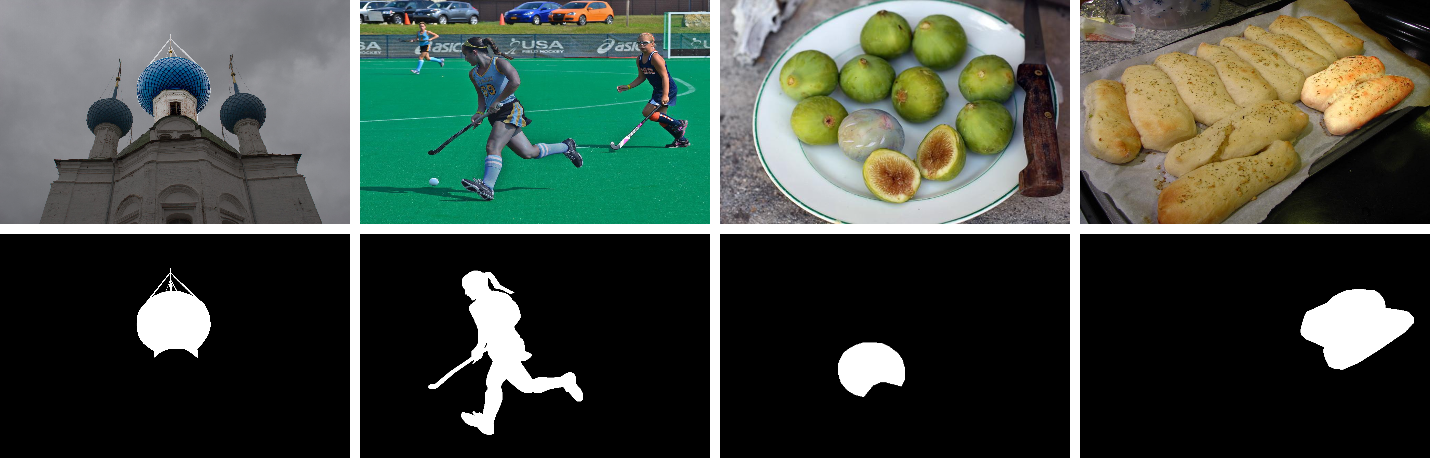}
    \caption{Examples of inharmonious images and the corresponding masks.}
    \label{Examples}
\end{figure}

 Following the definition in \cite{DIRL}, in a synthetic image, the region incompatible with the background in terms of color or illumination is named the \textit{inharmonious region}. Examples of synthetic images with inharmonious regions are shown in Figure~\ref{Examples}.
To avoid ambiguity, following \cite{DIRL}, we assume that the area of the inharmonious region is smaller than $50\%$ of the whole image. Otherwise, the background is viewed as the inharmonious region. 
The \textit{inharmonious
region localization} task aims to localize the inharmonious region. After the inharmonious region is localized, we can manually adjust the inharmonious region or utilize off-the-shelf image harmonization techniques to make the synthetic image more harmonious. Thus,  inharmonious  region  localization  task  is  necessary for image harmonization when the foreground mask is not available \cite{cun2020improving}. 
To the best of our knowledge, the only existing work on inharmonious region localization is DIRL \cite{DIRL}. However, it did not exploit the uniqueness of this task, \emph{i.e.}, the discrepancy between the inharmonious region and the background region.

Considering the uniqueness of the inharmonious region localization task, we treat it as a clustering problem and draw inspiration from typical clustering algorithm K-means \cite{kmeans}.  
Specifically, we aim to assign all the pixels into two clusters: inharmonious cluster and background cluster. The idea of K-means is iteratively performing the following two steps. 1) \emph{Assignment step}: for each pixel, calculate  the similarity  between  it and all the centroids  followed by assigning it to its nearest centroid. 2) \emph{Update step}: re-calculate the centroid of each cluster based on its associated samples. Inspired by K-means \cite{kmeans}, we design a Recurrent Self-Reasoning (RSR) module and insert it into the bottleneck of UNet \cite{Unet} structure. 

Now we briefly introduce our RSR module. We extract conventional feature map and style feature map from the encoder of UNet, in which the style feature map is expected to contain color and illumination information while there is no restriction for the conventional feature map.
Given the style feature map and an initial inharmonious mask, we calculate the background style feature by averaging  pixel-wise features within the background region. Then, we calculate a similarity map between the background style feature and the whole style feature map. This is similar to the \emph{assignment step} in K-means because the  similarity map roughly indicates which pixels belong to the background cluster. However, directly using this similarity map for assignment is inaccurate, because localizing the inharmonious region requires rich context information and the style feature may not be adequately informative. Therefore, we employ a convolutional Gate Recurrent Unit (GRU) cell \cite{GRU,convGRU}, which takes in the similarity map, current inharmonious mask, and conventional feature map to conduct self-reasoning about the location of inharmonious region and outputs an updated inharmonious mask. With a new inharmonious mask, we can update the background style feature, which is similar to re-calculating the centroids in the \emph{update step} of K-means. We repeat the above procedure iteratively and the quality of inharmonious mask is gradually improved. 

After introducing our core  Recurrent Self-Reasoning (RSR) module, we elaborate on our Recurrent Self-Reasoning Network (RSRNet).
As mentioned above, we insert our RSR module into the bottleneck of UNet structure. After the recurrent steps in RSR module, we feed the estimated mask from RSR to the decoder as attention guidance and the decoder outputs a refined mask. We observe that the mask from RSR can provide a compact shape of inharmonious region, while the  mask from decoder can provide sharper details and edges. Since these two masks are complementary to each other, we combine them adaptively to form our final mask. 
The main contributions of our work can be summarized
as follows:
\begin{itemize}
\item We treat inharmonious region localization as a clustering problem and draw inspiration from typical clustering algorithm, which provides a new perspective for this task. 
\item We propose a novel Recurrent Self-Reasoning (RSR) module to gradually improve the quality of inharmonious mask. We also propose to adaptively combine the  mask from RSR module and the mask from decoder, leading to further improvement.  
\item Extensive experiments on iHarmony4 dataset show that our RSRNet achieves the best performance both quantitatively and qualitatively.
\end{itemize}

\section{Related Work}
\subsection{Image Manipulation Localization}
Existing image manipulation localization methods can be categorized according to the type of manipulation (\emph{e.g.}, copy-move, removal, enhancement, and splicing), which share some similarities with inharmonious region localization task. Traditional methods are mainly aiming at localizing a
specific type of manipulation. These methods are generally based on the detection of specific clues or traces in the manipulated images including noise patterns \cite{mahdian2009using,noise2014exposing}, JPEG compression differences \cite{lin2009fast,li2009passive}, and colour filter array interpolation patterns \cite{dirik2009image,ferrara2012image}. Recently, there are many deep learning-based  methods \cite{MFCN,MantraNet,MAGritte,H-LSTM,Constrained-RCNN,GSRNet,SPAN}, which have shown impressive performance on image manipulation detection and  localization. However, these methods do not specifically consider the inconsistency due to color or illumination characteristics.

\subsection{Recurrent Neural Network}
Recurrent Neural Network (RNN) is a type of neural networks that recursively process sequential data. The Long Short-Term Memory (LSTM) \cite{lstm} and Gated Recurrent Unit (GRU) \cite{GRU} are two commonly used recurrent neural networks. Although these recurrent models are originally designed to solve machine translation and sequence modeling tasks, they are then widely employed to solve computer vision tasks, such as object recognition \cite{objectrecognition}, action detection \cite{actiondetection}, motion prediction\cite{motionprediction}, optical flow estimation \cite{RAFT}, and image manipulation localization \cite{H-LSTM}. The convolutional based GRU was proposed in \cite{convGRU}. Different from the above works, we are the first to apply RNN for the inharmonious region localization task.
\subsection{Image Harmonization and Inharmonious Region Localization}
Given a synthetic image which has incompatible foreground and background due to different color and illumination, image harmonization aims to adjust the appearance of the foreground to make it compatible with the background. Recently, abundant deep learning based image harmonization methods \cite{hao2020image,ling2021region,intriharm,bargainnet,cong2022high,hang2022scs,bao2022deep,CharmNet} have been proposed. To name a few, Tsai \emph{et al.} \cite{tsai2017dih} first proposed an end-to-end convolutional neural network to extract context and semantic information to generate the harmonized image. Cun and Pan \cite{cun2020improving} proposed  a spatial separated attention module S$^2$AM to separately learn the features in foreground and background. Cong \emph{et al.} \cite{DoveNet,bargainnet} introduced the concept of domain to distinguish incompatible regions and translate the foreground to the same domain as the background. Combing semantic features into the harmonization network  \cite{sofiiuk2021foreground} has shown great performance. Transformer based structure is utilized in  \cite{Guo_2021_ICCV} to tackle the harmonization task.
Most image harmonization methods require the inharmonious region mask as  input, which are usually hard to acquire in real-life scenario. Although S$^2$AM used the attention block to predict the inharmonious region mask, the quality of its predicted mask is far from satisfactory.
\par DIRL \cite{DIRL} is the first method focusing on inharmonious region localization, which effectively fused multi-scale features and suppressed redundant information to better localize the inharmonious region. However, it is a rather general model without exploiting the uniqueness of this task, that is, the discrepancy between the harmonious and inharmonious region.


\section{Methodology}
\subsection{Overview}
\begin{figure}[t]
    \centering
    \includegraphics[scale = 0.36]{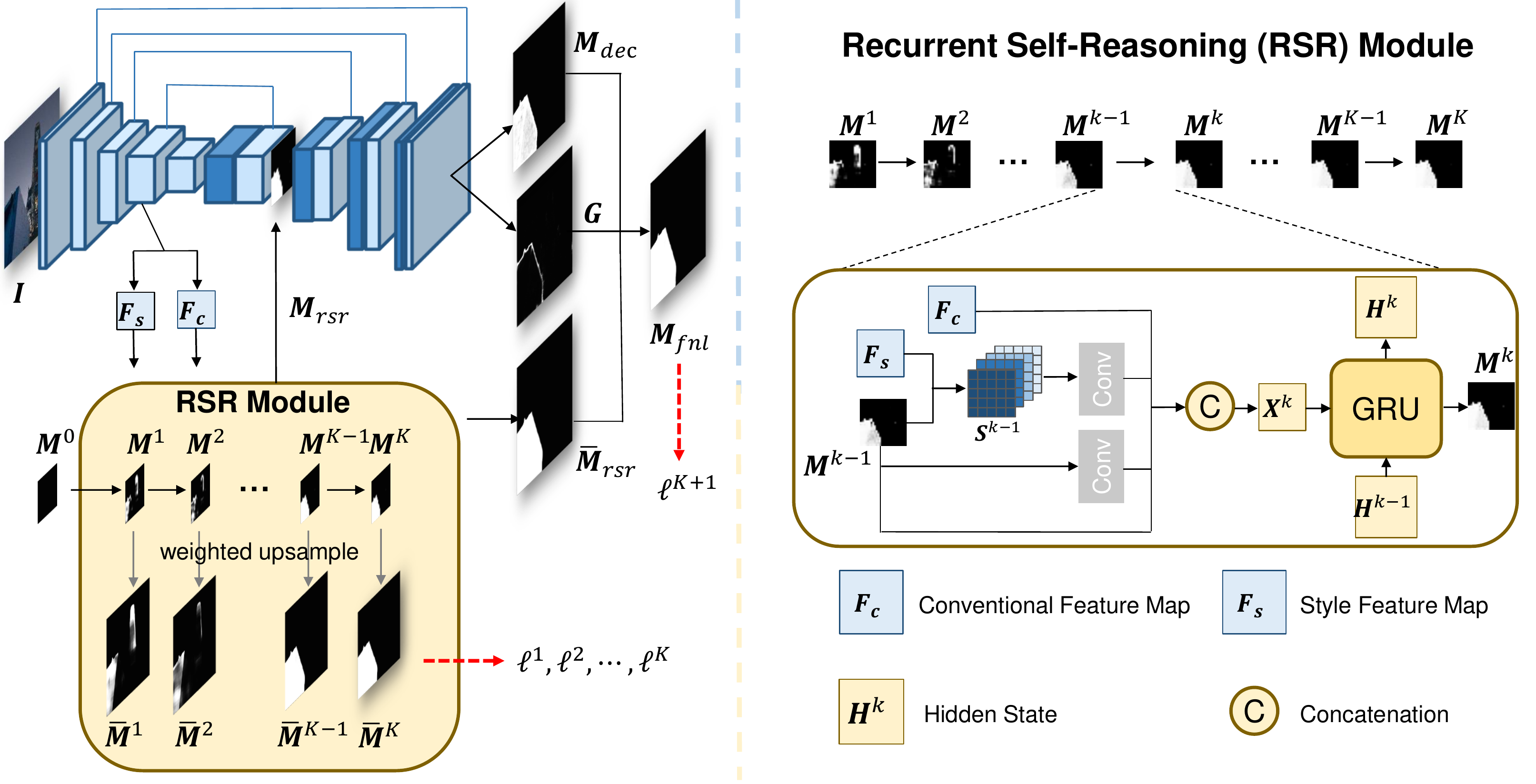}
    \caption{The left part shows the overall pipeline of our RSRNet which consists of a UNet structure and a RSR module inserted at the bottleneck. The right part shows the detailed iterative process in the RSR module.}
    
    \label{pipeline}
\end{figure}
Given an input synthetic image $\bm{I} \in \mathbb{R}^{H \times W \times 3}$, our goal is to estimate its inharmonious region mask $\bm{M} \in \mathbb{R}^{H \times W \times 1}$. Our whole model shown in Figure~\ref{pipeline} contains two parts: a UNet structure network and a Recurrent Self-Reasoning (RSR) module. The encoder contains five res-blocks. The first four res-blocks are adopted from ResNet34 \cite{ResNet} without pooling layer. After the fourth res-block, we get a feature map of shape $\frac{H}{8} \times \frac{W}{8} \times 512$. This feature map is then fed into two separate feature heads which both contain a $3 \times 3$ convolution followed by a Relu activation and a $1 \times 1$ convolution. These two feature heads extract a style feature map $\bm{F}_s$ and a conventional feature map $\bm{F}_c$ of shape $\frac{H}{8} \times \frac{W}{8} \times 256$, respectively. $\bm{F}_s$, $\bm{F}_c$, and the initial inharmonious mask $\bm{M}^{0}$ are delivered to Recurrent Self-Reasoning (RSR) module to refine the inharmonious mask iteratively. The mask $\bm{M}_{rsr}$ from RSR  is then provided for the decoder as attention guidance to output a refined mask $\bm{M}_{dec}$. The final output $\bm{M}_{fnl}$ will be the weighted combination of $\bm{M}_{dec}$ and upsampled $\bar{\bm{M}}_{rsr}$. Next, we will introduce our Recurrent Self-Reasoning (RSR) module in Section~\ref{sec:rsr} and the weighted combination of two masks in Section~\ref{sec:adapt_comb}.

\subsection{Recurrent Self-Reasoning Module (RSR)} \label{sec:rsr}
In this module, we will iteratively update the inharmonious region mask via self-reasoning. At first, the inharmonious mask is initialized as all zeros, \emph{i.e.}, $\bm{M}^{0}=\mathbf{0}$. We denote the estimated mask after the $k$-th iteration as $\bm{M}^{k}$. 

In the $k$-th iteration, we update the background style feature $\bm{f}^{k-1}_{bg}$, which is the average of style features within the background region based on the style feature map $\bm{F}_s$ and current inharmonious mask from last iteration $\bm{M}^{k-1}$. This step is similar to the \emph{update step} in K-Means algorithm where we update the centroid (the centroid is the background style feature representing the background harmonious centroid in our case) based on the current assignment of all samples (the assignment is the current estimated mask in our case).  A Multi-scale Similarity Map (MSM) $\bm{S}^{k-1}$ is constructed to measure how similar each pixel in $\bm{F}_s$ is to the background style feature $\bm{f}^{k-1}_{bg}$. Then, MSM $\bm{S}^{k-1}$, conventional feature $\bm{F}_c$, and inharmonious mask $\bm{M}^{k-1}$ will be passed through a GRU cell to produce an updated inharmonious mask $\bm{M}^{k}$. This corresponds to the \emph{assignment step} in K-Means where each sample is assigned to its corresponding centroid based their similarity. The detailed process of an iteration in RSR module is shown in the right part of Figure~\ref{pipeline}.

\subsubsection{Multi-scale Similarity Map}
In the $k$-th iteration, given the current inharmonious mask $\bm{M}^{k-1}$, we calculate the background style feature by averaging the style features within the background region.  Since the values of inharmonious mask are between 0 and 1 (1 for inharmonious and 0 for background), we set a threshold $\epsilon=0.5$ to determine pixels belonging to the background region. Then, the background style feature can be calculated as follows,  
\begin{eqnarray}
\bm{f}^{k-1}_{bg} = \frac{1}{\sum_i \delta(\bm{M}^{k-1}(i) < \epsilon)}\sum_i \delta(\bm{M}^{k-1}(i) < \epsilon) \bm{F}_{s}(i),
\end{eqnarray}
where $\bm{M}^{k-1}(i)$ (\emph{resp.}, $\bm{F}_{s}(i)$) is the value of $i$-th pixel in $\bm{M}^{k-1}$ (\emph{resp.}, $\bm{F}_{s}$). $\delta(s)$ is an indicator function, \emph{i.e.}, $\delta(s)=1$ if $s$ is true and $0$ otherwise. $\sum_i \delta(\bm{M}^{k-1}(i) < \epsilon)$ denotes the total number of background pixels. 

When calculating the similarity between $\bm{f}^{k-1}_{bg}$ and each pixel-level style feature, we consider the average feature in a local neighbourhood of each pixel, due to the potential noise of pixel-level features. Therefore, we use different neighborhood sizes to take multi-scale information into account. 
For scale $l$, we define  $\mathcal{N}^{l}(i)$ as a square neighborhood centered at the $i$-th pixel with side length $2l+1$. Then, we can calculate the cosine similarity between  $\bm{f}^{k-1}_{bg}$ and the local average feature at the $i$-th pixel:
\begin{eqnarray}
    \bm{S}^{k-1,l}(i) = \frac{\bm{f}^{k-1}_{bg} \cdot \sum_{j \in \mathcal{N}^l(i)} \bm{F}_s(j) }{ ||\bm{f}^{k-1}_{bg}||_2 \,\,||\sum_{j \in \mathcal{N}^{l}(i)}\bm{F}_s(j)||_2},
\end{eqnarray}
in which $\bm{S}^{k-1,l}(i)$ is the value of the $i$-th pixel in the similarity matrix $\bm{S}^{k-1,l}$ for scale $l$ in the $k$-th iteration, and $\cdot$ means the the dot product between two vectors.
In practice, we take a set of scales $l$ = 0, 1, 2, 3,  corresponding to the side length of neighborhood $2l+1$ = 1, 3, 5, 7, respectively. As a result, we obtain four similarity matrices, which are concatenated to form our Multi-scale Similarity Map (MSM) $\bm{S}^{k-1}$.

Note that the style feature map $\bm{F}_s$ will remain unchanged in the iterative process, but $\bm{S}^{k-1}$ will be changed because $\bm{f}^{k-1}_{bg}$ will be updated based on the current inharmonious mask $\bm{M}^k$ in each iteration. Recall that the initial $\bm{M}^0$ is an all-zero mask and thus $\bm{f}^0_{bg}$ is the global average feature over the entire image. Since the area of the inharmonious region is smaller than background region as defined in Section~\ref{sec:intro}, the initial $\bm{f}^0_{bg}$ is closer to the average feature of background region than that of inharmonious region. In the iterative process, $\bm{f}^{k-1}_{bg}$ will gradually approach the true average feature of the background region, offering guidance for inharmonious region localization in the subsequent iterations. 

\subsubsection{GRU Based Recurrent Unit}
In the $k$-th iteration, we have the current inharmonious mask $\bm{M}^{k-1}$, conventional feature map $\bm{F}_c$, MSM $\bm{S}^{k-1}$, and the hidden state from last iteration $\bm{H}^{k-1}$. $\bm{M}^{k-1}$ and $\bm{S}^{k-1}$ will go through two 3$\times$3 convolutions, respectively. The outputs will be concatenated with $\bm{F}_c$ and $\bm{M}^{k-1}$ to form the input $\bm{X}^k$. After that, $\bm{X}^k$ and  $\bm{H}^{k-1}$ are fed into a convolution GRU \cite{convGRU}. In each iteration, we learn the residual of the estimated mask $\Delta \bm{M}^k$, so $\bm{M}^{k} = \bm{M}^{k-1} + \Delta \bm{M}^k$. The update process in convolution GRU is
\begin{align*}
    \bm{Z}^k &= \sigma({\rm Conv}([\bm{H}^{k-1}, \bm{X}^k], \bm{W}_z)), \\
    \bm{R}^k &= \sigma({\rm Conv}([\bm{H}^{k-1}, \bm{X}^k], \bm{W}_r)), \\
    \tilde{\bm{H}^k} &= {\rm tanh}({\rm Conv}([\bm{R}^k \odot \bm{H}^{k-1}, \bm{X}^k], \bm{W}_h)), \\
    \bm{H}^k &= (1-\bm{Z}^k)\odot \bm{H}^{k-1} + \bm{Z}^k \odot \tilde{\bm{H}^k},
\end{align*}
in which ${\rm Conv}(\cdot, \bm{W})$ means $3\times 3$ convolution with kernel parameters $\bm{W}$, $[\cdot,\cdot]$ means concatenation, $\sigma$ is the sigmoid function, $\odot$ means element-wise product. $\bm{Z}$, $\bm{R}$, $\tilde{\bm{H}}$, and $\bm{H}$ are the latent variables defined in GRU \cite{GRU}.  $\Delta \bm{M}^k$ is obtained by applying a $3\times 3$ convolution followed by a $1\times1$ convolution on $\bm{H}^k$.

The updated mask $\bm{M}^{k}$ is of size $\frac{H}{8} \times \frac{W}{8}$. Similar to \cite{RAFT}, we use two convolutional layers to predict an upsample weight map of shape $\frac{H}{8}\times \frac{W}{8}\times(8\times8\times9)$ to upsample $\bm{M}^{k}$ to the full-resolution one $\bar{\bm{M}}^{k}$ by taking the weighted combination over the 9 neighborhood pixels. We update $\bm{M}^{k}$ at each iteration to $\bar{\bm{M}}^{k}$, so that the mask supervision can be employed at full resolution (see Section~\ref{sec:loss}).

\subsection{Adaptive Combination of Two Masks} \label{sec:adapt_comb}

We denote the mask from the last iteration in our RSR module as $\bm{M}_{rsr}$ and its upsampled version is $\bar{\bm{M}}_{rsr}$. $\bm{M}_{rsr}$ contains the general shape and location of the inharmonious region,
but lacks accurate edges and details (see examples in Supplementary) because the process is performed at the resolution of $\frac{1}{8}$.
We feed $\bm{M}_{rsr}$ into the decoder by concatenation to provide attention guidance. The decoder can focus on the inharmonious region provided by RSR module and integrate multi-scale encoder features through skip connection, producing the refined inharmonious mask $\bm{M}_{dec}$ with sharper details and edges. 

In our experiments, we observe that although $\bm{M}_{dec}$ is more accurate about edges and details, it may contain some holes and misdetected regions (see examples in Supplementary). Thus, $\bm{M}_{dec}$ and $\bar{\bm{M}}_{rsr}$ are complementary to each other and should be utilized simultaneously. 
For two complementary outputs, it is more beneficial to adaptively combine them  instead of simply calculating the average \cite{gatedfusion,CDFI}. 
Thus, we learn a combination mask $\bm{G}$ with values between 0 and 1 to adaptively combine $\bm{M}_{dec}$ and $\bar{\bm{M}}_{rsr}$. 
Then the final mask $\bm{M}_{fnl}$ can be obtained by
\begin{equation*}
    \bm{M}_{fnl} = \bm{G} \odot \bm{M}_{dec} + (\bm{1}-\bm{G}) \odot \bar{\bm{M}}_{rsr}.
\end{equation*}
The combination mask is expected to utilize the accurate details and edges from $\bm{M}_{dec}$ as well as the general shape and location from $\bar{\bm{M}}_{rsr}$ to form our final mask. 

\subsection{Loss Function} \label{sec:loss}
In our RSR module, we perform $K$ iterations and 
obtain $K$ upsampled masks $\bar{\bm{M}}^k$ for $k=1,\ldots, K$. We choose $K=12$ by cross-validation.
We denote the loss for the $k$-th mask as $\ell^{(k)}$. Besides, we denote the loss for the final mask $\bm{M}_{fnl}$ as $\ell^{(K\!+\!1)}$.
Given the ground-truth mask $\bm{M}$ and an estimated mask $\bar{\bm{M}}^k$, the loss $\ell^{(k)}$ is comprised of three parts following \cite{Basnet}: $\ell^{(k)} = \ell^{(k)}_{bce} + \ell^{(k)}_{ssim} + \ell^{(k)}_{iou}.$

$l^{(k)}_{bce}$ is the binary cross-entropy loss, which is commonly used in binary classification task and segmentation task. 
$\ell^{(k)}_{ssim}$ is the structural similarity loss, which can better represent the structural information of the ground-truth mask.
$\ell^{(k)}_{iou}$ is the Intersection over Union (IoU) loss.
We refer readers to \cite{Basnet} for details about these three losses. Our total loss is a weighted sum of losses for all the estimated masks:
\begin{eqnarray} \label{eqn:L_total}
    \mathcal{L} = \ell^{(K\!+\!1)} + \sum_{k = 1}^{K} \lambda^{K-k}\ell^{(k)},
\end{eqnarray}
where $\lambda$ is a weight factor set as 0.8 following \cite{RAFT}. The loss weights for the masks $\bar{\bm{M}}^k$ are increasing exponentially since the mask is initially coarse and gradually getting more accurate, and we assign higher weights to the more accurate masks.

\section{Experiments}
\subsection{Dataset and Evaluation Metrics}
Following \cite{DIRL}, we conduct our experiments on the the image harmonization dataset \textbf{iHarmony4}  \cite{DoveNet}, which contains inharmonious images with corresponding masks and harmonious images. iHarmony4 \cite{DoveNet} consists of four sub-datasets: HCOCO, HFlickr, HAdobe5K, and Hday2night. The iHarmony4 dataset is suitable for inharmonious region localization task since the inconsistency between foreground and background is mainly due to incompatible color or illumination \cite{DIRL}.
For HCOCO and HFlickr sub-datasets, the inharmonious images are  generated by applying color transfer methods \cite{reinhard2001color,xiao2006color,fecker2008histogram,pitie2007automated} to transfer the foreground color of real images. For HAdobe5K sub-dataset, real images are retouched by five professional photographers to obtain the corresponding inharmonious images. For Hday2night sub-dataset, the inharmonious image is made by overlaying the foreground of a real image with the corresponding region in another image, which has the same scene captured under a different condition.  To avoid the ambiguity of the definition of inharmonious region, we follow \cite{DIRL} to only keep the images with foreground area larger than $50\%$. We follow the same train-test split as \cite{DIRL} to get 64255 training images and 7237 test images. Following \cite{DIRL}, we adopt the evaluation metrics including  Average Precision (AP), $F_1$ score, and Intersection over Union (IoU). 

\subsection{Implementation Details}
We adopt the first four blocks from ImageNet-pretrained ResNet34 \cite{ResNet} as the first four blocks in our encoder, and the last block in our encoder is similar to the first block. 
We also use 3$\times$3 convolution to replace the 7$\times$7 convolution and remove the pooling layer to keep the resolution and retain more details in the shallow layer. 
We use Pytorch \cite{pytorch} to implement our model. We use Adam optimizer with $\beta_1 = 0.9$, $\beta_2 = 0.999$, weight decay = 1e-4, and initial learning rate = 1e-4. We train our model with batch size 32 for 60 epochs on a  Ubuntu 18.04.4 machine with 4 GeForce GTX TITAN X GPUs with the learning rate reduced by 0.5 after 30, 40, 50, 55 epochs respectively. The random seed set for numpy and Pytorch is 42.

\subsection{Comparison with the State-of-the-art}
To the best of our knowledge, DIRL \cite{DIRL} is the only work focusing on the inharmonious region localization task. Therefore, following \cite{DIRL}, we also compare our model with the state-of-the-art methods from other related fields. We choose three groups of baselines for comparison: 1) Popular segmentation networks: UNet \cite{Unet}, DeepLabv3 \cite{DeepLabv3}, HRNet-OCR \cite{HRNet}, SegFormer \cite{xie2021segformer}. 2) Image manipulation localization methods: MFCN \cite{MFCN}, MantraNet \cite{MantraNet}, MAGritte \cite{MAGritte}, H-LSTM \cite{H-LSTM}, SPAN \cite{SPAN}. 3) Salient object detection methods: F3Net \cite{F3net}, GATENet \cite{GATENet}, MINet \cite{MINet}

\subsubsection{Quantitative Evaluation}
We report AP, $F_1$, and IoU of all methods for each sub-datasets in Table~\ref{result}. Following \cite{DIRL}, we use ResNet34 as backbone for ResNet based methods, HRNet30 for HRNet-OCR, and SegFormer-B3 for SegFormer. Based on Table~\ref{result}, our method achieves the best performance on the whole dataset. Our method  beats the best method MAGritte in the image manipulation localization field by a large margin. The general segmentation methods and MINet in the salient object detection field also show competitive performance, but they are still worse than our model. For the strongest baseline DIRL, our model has 3.19$\%$ (\emph{resp.}, 3.96$\%$, 5.57$\%$) improvement for
AP, $F_1$, and IoU.

In addition, we report model related statistics including amount of parameters, model inference speed, and GFlops of our model and other strong baselines in the Supplementary.

\begin{table}[t]
\centering
\setlength{\tabcolsep}{1mm}

\scalebox{0.70}{
\begin{tabular}{cccccccccccccccc}
\hline
\multirow{2}{*}{Method} & \multicolumn{3}{c|}{HCOCO}                                                          & \multicolumn{3}{c|}{HAdobe5k}                                                      & \multicolumn{3}{c|}{HFlickr}                                                        & \multicolumn{3}{c|}{Hday2night}                                                     & \multicolumn{3}{c}{\textbf{All}}                                                   \\ \cline{2-16} 
                        & \multicolumn{1}{c|}{AP$\uparrow$ }   & \multicolumn{1}{c|}{$F_1\uparrow$ }    & IoU$\uparrow$                         & \multicolumn{1}{c|}{AP$\uparrow$ }  & \multicolumn{1}{c|}{$F_1\uparrow$ }    & IoU$\uparrow$                         & \multicolumn{1}{c|}{AP$\uparrow$ }   & \multicolumn{1}{c|}{$F_1\uparrow$ }    & IoU$\uparrow$                         & \multicolumn{1}{c|}{AP$\uparrow$ }   & \multicolumn{1}{c|}{$F_1\uparrow$ }    & IoU$\uparrow$                         & \multicolumn{1}{c|}{AP$\uparrow$ }   & \multicolumn{1}{c|}{$F_1\uparrow$ }    & IoU$\uparrow$                        \\ \hline
UNet                   & 68.11                     & 0.5869                     & 56.57                      & 89.26                    & 0.8380                     & 80.85                      & 80.72                     & 0.7683                     & 74.58                      & 35.74                     & 0.2362                     & 19.60                      & 74.90                     & 0.6717                     & 64.74                     \\
DeepLabv3               & 69.09                     & 0.6070                     & 58.21                      & 90.20                    & 0.8591                     & 81.56                      & 80.01                     & 0.7698                     & 74.91                      & 35.87                     & 0.2550                     & 21.38                      & 75.69                     & 0.6902                     & 66.01                     \\
HRNet-OCR               & 68.89                     & 0.5981                     & 57.69                      & 89.63                    & 0.8387                     & 80.98                      & 79.62                     & 0.7489                     & 74.55                      & 34.98                     & 0.2477                     & 21.34                      & 75.33                     & 0.6765                     & 65.49                     \\
SegFormer               & 72.46                     & 0.6578                     & 58.78                      & 89.43                    & 0.8531                     & 80.44                      &  {85.19}                     &  {0.7986}                     &  {75.02}                      &  {45.16}                     &  {0.3856}                     &  {32.75}                      & 78.05                     & {0.7249}                     & {66.55}                     \\ \hline
MFCN                    & 37.36                     & 0.3030                     & 25.18                      & 62.75                    & 0.5365                     & 36.63                      & 49.89                     & 0.4209                     & 28.34                      & 19.71                     & 0.1426                     & 11.88                      & 45.63                     & 0.3794                     & 28.54                     \\
MantraNet               & 56.55                     & 0.4811                     & 41.04                      & 81.07                    & 0.7510                     & 68.50                      & 67.52                     & 0.6302                     & 58.51                      & 28.88                     & 0.2019                     & 16.71                      & 64.22                     & 0.5691                     & 50.31                     \\
MAGritte                & 64.75                     & 0.6058                     & 51.77                      & 85.50                    & 0.8630                     & 76.36                      & 75.02                     & 0.7725                     & 70.25                      & 31.20                     & 0.2549                     & 17.05                      & 71.16                     & 0.6907                     & 60.14                     \\
H-LSTM                  & 52.29                     & 0.4336                     & 37.81                      & 77.62                    & 0.7056                     & 65.19                      & 63.12                     & 0.5918                     & 54.93                      & 26.59                     & 0.1977                     & 15.91                      & 60.21                     & 0.5239                     & 47.07                     \\
SPAN                    & 58.41                     & 0.4906                     & 45.07                      & 82.57                    & 0.7786                     & 72.49                      & 69.22                     & 0.6510                     & 62.20                      & 29.58                     & 0.2171                     & 19.41                      & 65.94                     & 0.5850                     & 54.27                     \\ \hline
F3Net                   & 54.17                     & 0.4703                     & 40.03                      & 74.31                    & 0.6944                     & 60.08                      & 72.53                     & 0.6582                     & 59.31                      & 30.08                     & 0.2563                     & 20.83                      & 61.46                     & 0.5506                     & 47.48                     \\
GATENet                 & 55.07                     & 0.4568                     & 38.89                      & 75.19                    & 0.6634                     & 59.18                      & 74.13                     & 0.6256                     & 57.51                      & 30.98                     & 0.2174                     & 19.38                      & 62.43                     & 0.5296                     & 46.33                     \\
MINet                   & 71.74                     & 0.6022                     & 55.79                      & 89.58                    & 0.8379                     & 77.23                      & 83.86                     & 0.7761                     & 72.51                      & 37.82                     & 0.2710                     & 19.38                      & 77.51                     & 0.6822                     & 63.04                     \\ \hline
DIRL                    &  {74.25}                     &  {0.6701}                     &  {60.85}                      &  {92.16}                    &  {0.8801}                     &  {84.02}                      & 84.21                     & 0.7786                     & 73.21                      & 38.74                     & 0.2396                     & 20.11                      &  {80.02}                     &  {0.7317}                     &  {67.85}                     \\
RSRNet                  &  \textbf{78.42} &  \textbf{0.7131} &  \textbf{65.85} &  \textbf{93.10} &  \textbf{0.8901} &  \textbf{85.96} &  \textbf{87.11} &  \textbf{0.8048} &  \textbf{76.84} &  \textbf{47.34} &  \textbf{0.3028} &  \textbf{26.34} &  \textbf{82.57} &  \textbf{0.7607} &  \textbf{71.63} \\ \hline
\end{tabular}}
\caption{Performance comparison with different methods on the iHarmony4
dataset. $\uparrow$ means the larger, the better. The best results are
denoted in  \textbf{bold}.}

\label{result}
\end{table}

\subsubsection{Qualitative Evaluation}
\begin{figure}[t]
    \centering
    \includegraphics[scale = 0.51]{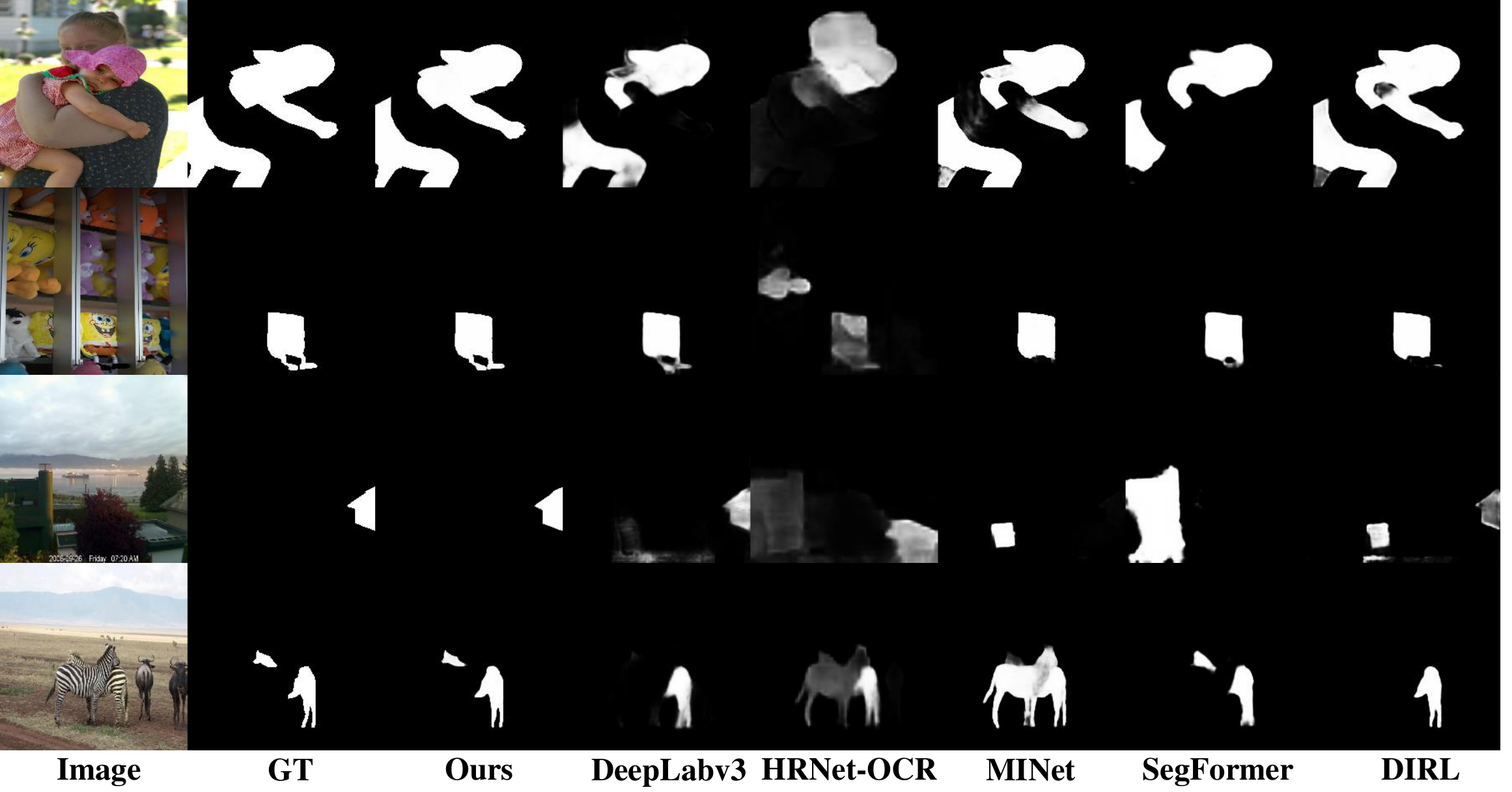}
    \caption{Qualitative comparison of our model with other state-of-the-art methods from related fields. GT is the ground-truth inharmonious mask.}
    \label{comparisonwithsota}
\end{figure}

To better verify the advantage of our model, we visualize the predicted masks from our model and top five baseline methods in Figure~\ref{comparisonwithsota}. It can be seen that in various challenging scenarios including occlusion, fine-grained structure, and small object, our RSRNet can better localize the inharmonious region with clear and sharp boundary.

\subsection{Ablation Studies}\label{sec:ablate}
\par Row 1 in Table~\ref{Abalation} is the basic UNet including an encoder and decoder. The result is slightly different from UNet in Table~\ref{result} because the original UNet method only has $\ell_{bce}$. 
In row 2, we insert our RSR module into the bottleneck and evaluate the mask $\bm{M}_{dec}$ output from decoder. Comparing row 1 \& 2,  the quality of $\bm{M}_{dec}$ is significantly improved after inserting RSR module. In row 3, we only iteratively update similarity matrix without using other information, which is closer to K-means clustering. Specifically, we directly use a single similarity matrix with scale $l=0$ as the updated inharmonious mask and update it for 12 iterations. The results demonstrate that the similarity map is not informative enough  and imposing supervision on it may degrade the performance. In row 4, we simply average $\bar{\bm{M}}_{rsr}$ and $\bm{M}_{dec}$ as $\bm{M}_{fnl}$, which is better than row 2 but much worse than our full method in row 9. The results show that it is necessary to combine $\bar{\bm{M}}_{rsr}$ and $\bm{M}_{dec}$ and sophisticated combination strategy performs more favorably. 

In row 5, we remove the GRU cell and only use two convolution layers to output the mask.
In row 6, we remove MSM, and only feed $\bm{F}_c$ and $\bm{M}^{k-1}$ into the GRU cell.
In row 7, we replace $\bm{F}_c$ with $\bm{F}_s$.
In row 8, we use bilinear upsample instead of the weighted upsample for $\bm{M}^k$. The results in row 5-8 are all worse than our full method, which justifies the necessity and effectiveness of our RSR module.

For each experiment from row 5 to row 9 in Table~\ref{Abalation}, we further report the results of $\bar{\bm{M}}_{rsr}$ and $\bm{M}_{dec}$  to be compared with $\bm{M}_{fnl}$ in Table~\ref{RSR abalation}. We can see that $\bar{\bm{M}}_{rsr}$ usually has a higher AP while $\bm{M}_{dec}$ has higher $F_1$ and IoU. The reason is that the AP is an average value computed at different thresholds while $F_1$ and IoU are both at only one threshold 0.5, and those uncertain or misdetected region in $\bm{M}_{dec} $ will have worse performance when the threshold is high. Combining these two complementary masks leads to better performance of $\bm{M}_{fnl}$. 

\begin{table*}[t]
\centering
\setlength{\tabcolsep}{1mm}
\scalebox{0.85}{
\begin{tabular}{|c|c|c|c|c|c|c|}
\hline
\multirow{2}{*}{\#} & \multicolumn{1}{|c|}{\multirow{2}{*}{UNet}} & \multicolumn{1}{c|}{\multirow{2}{*}{RSR}} & \multicolumn{1}{c|}{\multirow{2}{*}{Mask Combination}} & \multicolumn{3}{c|}{Evaluation} \\ \cline{5-7} 
& \multicolumn{1}{c|}{}                                         & \multicolumn{1}{c|}{}                     & \multicolumn{1}{c|}{}& AP($\%$)$\uparrow$       & $F_1\uparrow$        & IoU($\%$)$\uparrow$     \\ \hline
1& + & & & 78.35 & 0.7130 & 65.99 \\ \hline
2&+  & + & & 80.53 & 0.7426 & 69.72 \\ \hline
3&+ & only similarity map &  &77.24 &0.6973 &64.55 \\ \hline
4&+  & + & simple average &80.82  &0.7431  &69.69  \\ \hline
5&+  & + (w/o GRU) & + &81.66 &0.7564 &70.52 \\ \hline
6&+  & + (w/o MSM) & + &80.62 &0.7451 &69.81 \\ \hline
7&+ & + (w/o $\bm{F}_c$) & + & 81.08 &0.7440 &69.93\\ \hline
8&+ &+ (w/o weighted upsample) & + & 81.24 & 0.7515 & 70.57  \\ \hline
9&+  & + & + & \textbf{82.57} & \textbf{0.7607}    & \textbf{71.63} \\ \hline
\end{tabular}}
\caption{Ablation study on different components and combinations in our method. See Section~\ref{sec:ablate} for the detailed explanation.}

\label{Abalation}
\end{table*}

\begin{table}[t]
\centering
\setlength{\tabcolsep}{1mm}

\scalebox{0.80}{
\begin{tabular}{c|c|ccc|ccc|ccc}
\hline
\multirow{2}{*}{\#} & \multirow{2}{*}{RSR}  & \multicolumn{3}{c|}{$\bar{\bm{M}}_{rsr}$}                                                          & \multicolumn{3}{c|}{$\bm{M}_{dec}$}                                                                & \multicolumn{3}{c}{$\bm{M}_{fnl}$}                                                                   \\ \cline{3-11} 
                    &                       & \multicolumn{1}{c|}{AP($\%$)$\uparrow$} & \multicolumn{1}{c|}{$F_1\uparrow$} & IoU($\%$)$\uparrow$ & \multicolumn{1}{c|}{AP($\%$)$\uparrow$} & \multicolumn{1}{c|}{$F_1\uparrow$} & IoU($\%$)$\uparrow$ & \multicolumn{1}{c|}{AP($\%$)$\uparrow$} & \multicolumn{1}{c|}{$F_1\uparrow$}   & IoU($\%$)$\uparrow$ \\ \hline
1                   & w/o GRU               & \multicolumn{1}{c|}{81.29}              & \multicolumn{1}{c|}{0.7495}        & 70.21               & \multicolumn{1}{c|}{77.88}              & \multicolumn{1}{c|}{0.7526}        & 70.38               & \multicolumn{1}{c|}{81.66}              & \multicolumn{1}{c|}{0.7564}          & 70.52               \\
2                   & w/o MSM               & \multicolumn{1}{c|}{80.27}              & \multicolumn{1}{c|}{0.7398}        & 68.92               & \multicolumn{1}{c|}{75.57}              & \multicolumn{1}{c|}{0.7400}        & 69.15               & \multicolumn{1}{c|}{80.62}              & \multicolumn{1}{c|}{0.7451}          & 69.81               \\
3                   & w/o $\bm{F}_c$        & \multicolumn{1}{c|}{80.74}              & \multicolumn{1}{c|}{0.7392}        & 69.11               & \multicolumn{1}{c|}{76.79}              & \multicolumn{1}{c|}{0.7413}        & 69.58               & \multicolumn{1}{c|}{81.08}              & \multicolumn{1}{c|}{0.7440}          & 69.93               \\
4                   & w/o weighted upsample & \multicolumn{1}{c|}{80.57}              & \multicolumn{1}{c|}{0.7423}        & 69.26               & \multicolumn{1}{c|}{76.81}              & \multicolumn{1}{c|}{0.7486}        & 70.20               & \multicolumn{1}{c|}{81.24}              & \multicolumn{1}{c|}{0.7517}          & 70.57               \\
5                   & full module           & \multicolumn{1}{c|}{82.19}              & \multicolumn{1}{c|}{0.7547}        & 70.62               & \multicolumn{1}{c|}{78.32}              & \multicolumn{1}{c|}{0.7591}        & 71.42               & \multicolumn{1}{c|}{\textbf{82.57}}     & \multicolumn{1}{c|}{\textbf{0.7607}} & \textbf{71.63}      \\ \hline
\end{tabular}}
\caption{The evaluation results of three masks $\bar{\bm{M}}_{rsr}$, $\bm{M}_{dec}$, $\bm{M}_{fnl}$ using ablated RSR module.}
\label{RSR abalation}
\end{table}

\section{Conclusion}
In this paper, we propose a Recurrent Self-Reasoning based Network (RSRNet) to achieve inharmonious region localization. Inspired by K-means algorithm, we design a recurrent module to iteratively reason about the inharmonious region. We also design an adaptive combination mask to selectively combine the mask from our RSR module and that from the decoder. Experiments on iHarmony4  demonstrate the superiority of our proposed model.

\section*{Acknowledgement}

The work was supported by the Shanghai Municipal Science and Technology Major/Key Project, China (2021SHZDZX0102, 20511100300) and  National Natural Science Foundation of China (Grant No. 61902247).

\bibliography{egbib}
\end{document}


\maketitle

In this supplementary, we conduct comparison of computational complexity in Sec.~\ref{sec:model_size}, and study the effects of different loss components in Sec.~\ref{sec:loss_effects}. We provide  visualization results in Sec.~\ref{sec:visualization}. At last, we discuss our limitations in Sec.~\ref{sec:Limitation}.

\section{Computational Complexity Comparison} \label{sec:model_size}

We report the statistics related to computational complexity in Table~\ref{table:model_size}. We first compare the whole RSRNet with our RSR module. It can be seen that our RSR module is rather light-weighted and only adding small overhead to the backbone. We also compare with baselines DIRL, SegFormer, and MiNet, which shows that the computational complexity of our method is comparable with baselines. 

\begin{table}[t]
\centering

\begin{tabular}{c|c|c|c|c|c}
                     & RSR module & RSRNet & DIRL    & SegFormer & MINet   \\ \hline
Number of parameters & 5.52M      & 54.28M & 53.46M  & 48.28M    & 68.28M  \\ \hline
Inference time       & 52.8ms     & 90.6ms & 63.80ms & 61.36ms   & 63.95ms \\ \hline
GFlops               & 4.33       & 99.27  & 104.67  & 30.97     & 116.01 
\end{tabular}
\caption{Comparison between model size and speed. All the tests run on a single GeForce GTX TITAN X GPU.}
\label{table:model_size}
\end{table}

\section{Effects of Loss Functions} \label{sec:loss_effects}
We investigate the effectiveness of different loss terms $\ell_{bce}$, $\ell_{ssim}$, and $\ell_{iou}$ in Table~\ref{loss}. We can see that $\ell_{bce}$ is essential for our model, and the training process can not converge without it. We also observe that $\ell_{ssim}$ and $\ell_{iou}$ are both beneficial for the overall performance. We also set $\lambda = 1$ so that the weights for all the masks are the same. The results show that it is beneficial and reasonable to set exponentially increasing weights from coarse to fine outputs.

\begin{table}[t]

\centering
\begin{tabular}{c|cccc}
\hline
& \multicolumn{4}{c}{Loss} \\ \hline
Evaluation & \multicolumn{1}{c|}{w/o $\ell_{bce}$} & \multicolumn{1}{c|}{w/o $\ell_{ssim}$} & \multicolumn{1}{c|}{w/o $\ell_{iou}$} & $\lambda = 1$ \\ \hline
AP($\%$)$\uparrow$        & \multicolumn{1}{c|}{42.82}            & \multicolumn{1}{c|}{81.59}             & \multicolumn{1}{c|}{82.10}            & 81.38         \\
$F_1\uparrow$       & \multicolumn{1}{c|}{0.4255}           & \multicolumn{1}{c|}{0.7444}            & \multicolumn{1}{c|}{0.7524}           & 0.7529        \\
IoU($\%$)$\uparrow$        & \multicolumn{1}{c|}{35.34}            & \multicolumn{1}{c|}{69.95}             & \multicolumn{1}{c|}{70.37}            & 69.93     \\ \hline   
\end{tabular}
\caption{Ablation study on different loss terms. $\lambda$ is the hyper-parameter in Eqn. (3) in the main paper.}
\label{loss}
\end{table}

\section{Visualization Results} \label{sec:visualization}

First, we visualize the inharmonious region mask $\bar{\bm{M}}^k$ and the similarity matrix $\bm{S}^{k-1,l}$ ($l=0,3$) in each iteration in our RSR module. 

\begin{figure}[t]
    \centering
    \includegraphics[scale = 0.54]{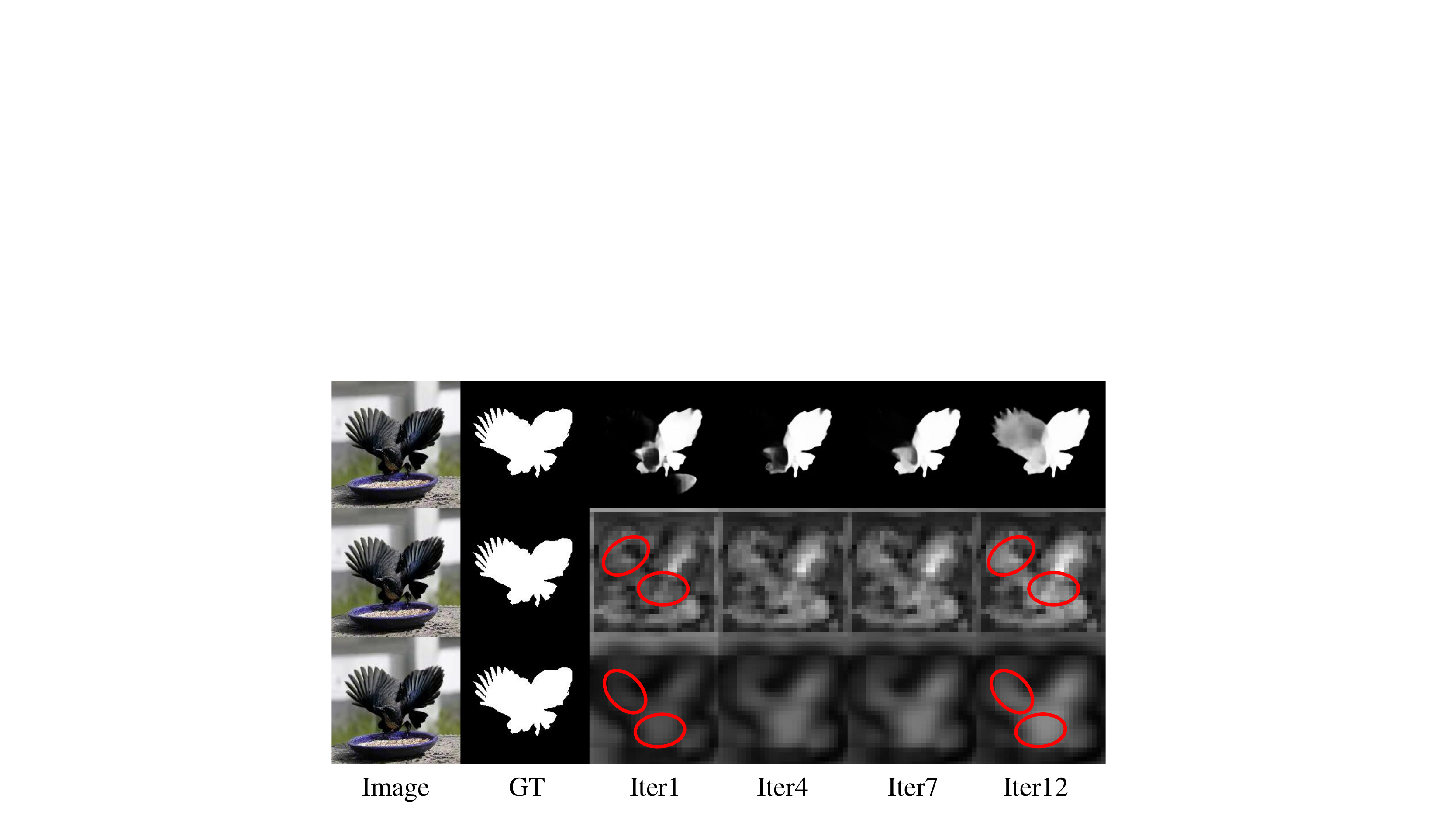}
    \caption{The update process of our RSR module at different iterations. At $k$-th iteration, we visualize the inharmonious region mask $\bar{\bm{M}}^k$ in the first row and similarity map $\bm{S}^{k-1,0}$ (\emph{resp.}, $\bm{S}^{k-1,3}$) with $l=0$ (\emph{resp.}, $l=3$) in the second (\emph{resp.}, third) row (areas with large changes are highlighted by red circles).  GT is the ground-truth inharmonious mask.}
    \label{iteration}
\end{figure}

 \begin{figure}[t]
     \centering
     \includegraphics[scale=0.45]{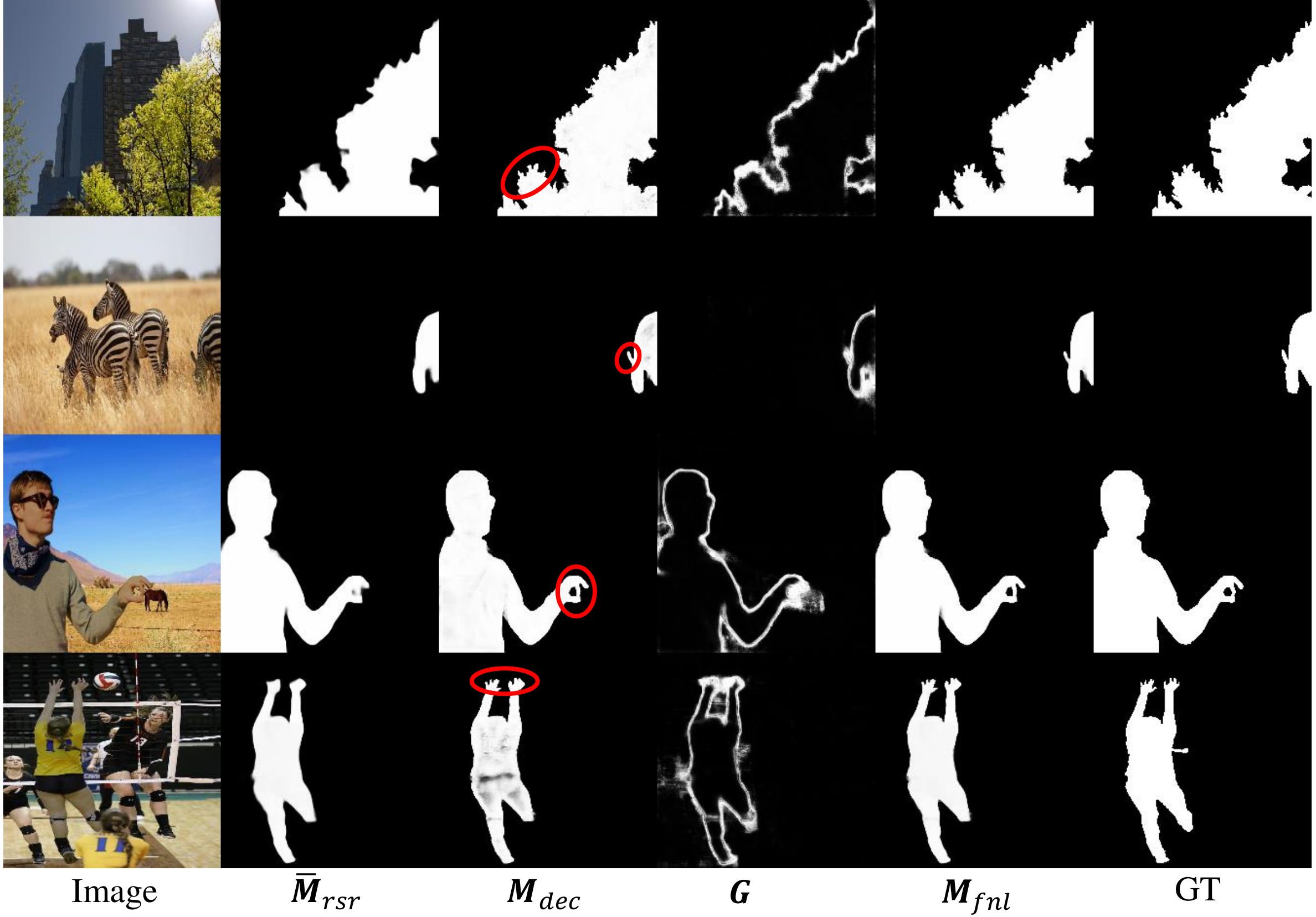}
     \caption{Visualization of three masks $\bar{\bm{M}}_{rsr}$, $\bm{M}_{dec}$, $\bm{M}_{fnl}$ and the combination mask $\bm{G}$. GT is the ground-truth inharmonious mask. Zoom in to see the edges and details which are highlighted by red circles. }
     \label{different_mask}
 \end{figure}
From Figure~\ref{iteration}, we can see that the inharmonious mask and the similarity matrix at the early stage miss part of inharmonious region and miss-classify some background region as inharmonious.
As the iteration goes, the detected inharmonious region is gradually recovered with higher confidence, with the similarity map providing more complete and accurate information.

When comparing similarity map and inharmonious mask, similarity map is able to provide the general location of inharmonious region but rather coarse, so we need a GRU cell to further process it. By comparing the similarity map of different scales ($l=0,3$), the one of smaller scale provides more complete yet more noisy inharmonious region, while the the one of larger scale is more conservative but less noisy.


Then, we visualize the masks $\bar{\bm{M}}_{rsr}$, $\bm{M}_{dec}$, $\bm{M}_{fnl}$ and the adaptive combination map $\bm{G}$ in Figure~\ref{different_mask}.
The goal of our adaptively
combined mask is utilizing the advantages of both $\bar{\bm{M}}_{rsr}$  and $\bm{M}_{dec}$. We can see that the mask $\bar{\bm{M}}_{rsr}$ from our RSR module is more confident about the general shape and location of the inharmonious object, so it is more compact without holes or uncertain areas in the inharmonious region. Nevertheless, it is less accurate with the edges and lacking in many details. For the mask $\bm{M}_{dec}$ from decoder, it has utilized more information from multi-scale encoder features. Hence, it can better segment the inharmonious region with sharp and accurate edges, for example, tree branches in row 1 and the zebra ear in row 2. 
    However, it may contain some holes or uncertain areas in the mask, like the human body in row 4. The adaptively combined mask $\bm{M}_{fnl}$ generally chooses the edges of the inharmonious region from $\bm{M}_{dec}$ and the inner part from $\bar{\bm{M}}_{rsr}$, which can be seen from the combination mask $\bm{G}$. Therefore, the combined mask $\bm{M}_{fnl}$  can have a compact mask without holes while keep the detailed and sharp edge information at the same time.

\begin{figure}[t]
    \centering
    \includegraphics[scale = 0.94]{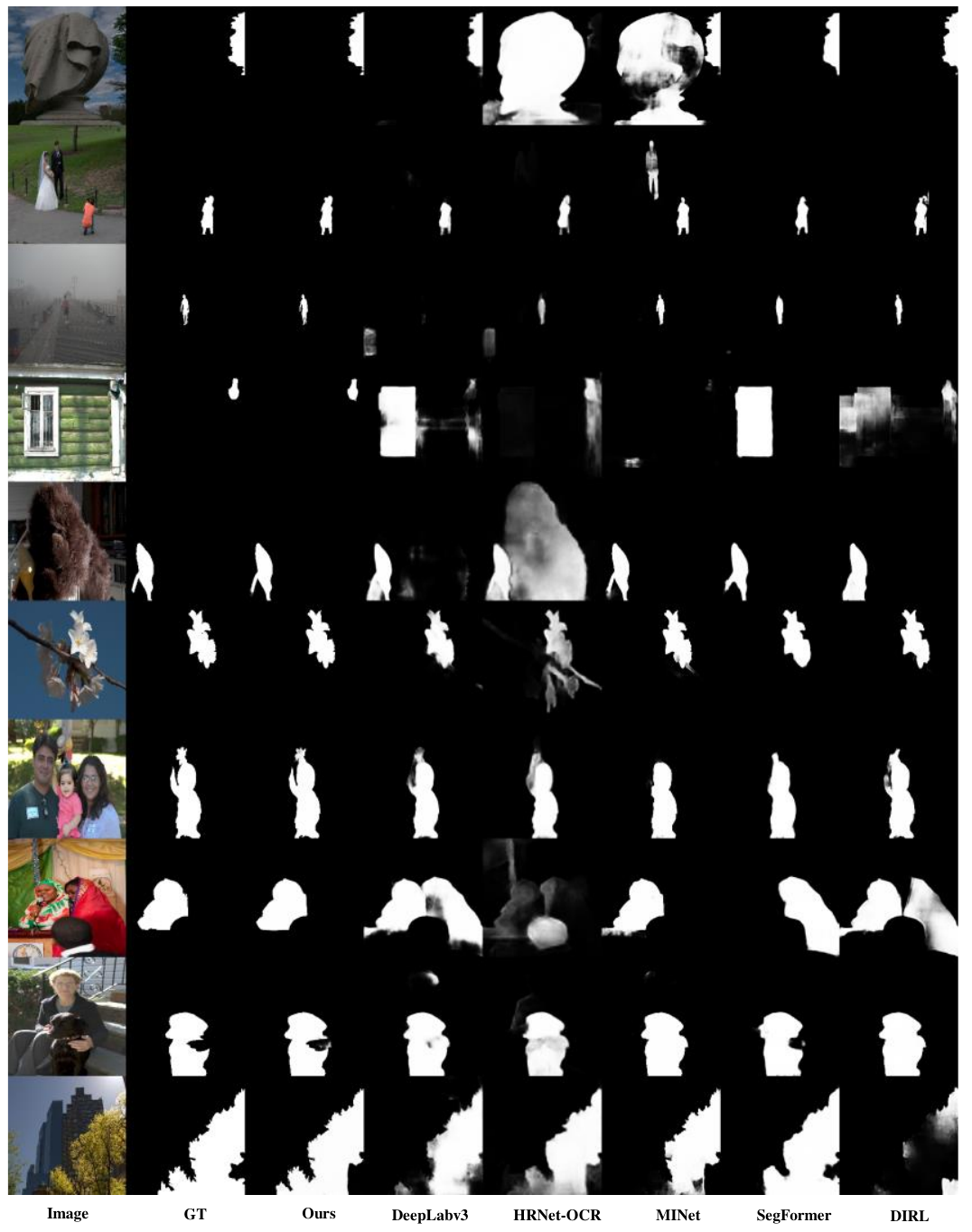}
    \caption{More qualitative  comparison  of  our  model  with  other  state-of-the-art  methods  from related fields. GT is the ground-truth inharmonious mask.}
    \label{fig:more baseline}
\end{figure}
We also provide more qualitative comparisons between our model and other baselines in Figure~\ref{fig:more baseline}.


\section{Limitation}\label{sec:Limitation}
\begin{figure}[t]
     \centering
     \includegraphics[scale=0.35]{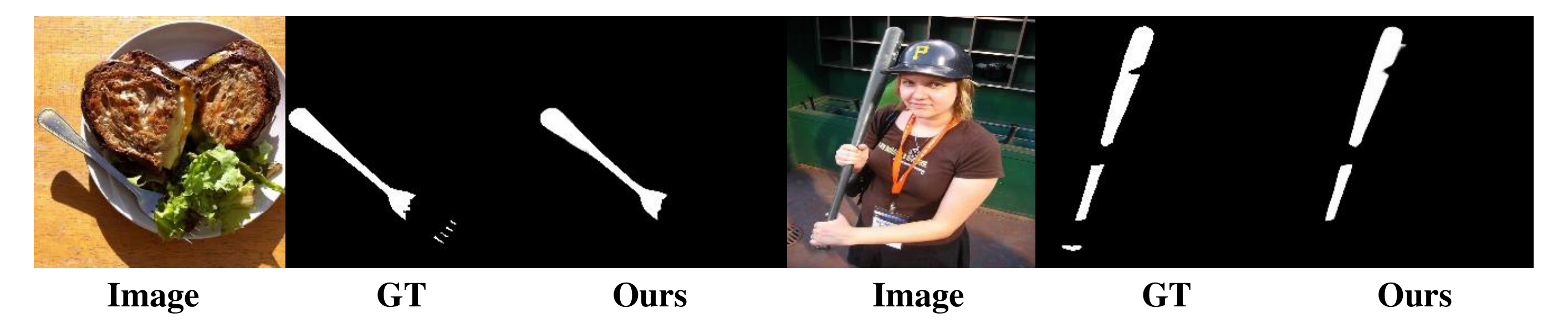}
     \caption{Our method only detects part of the inharmonious region in a few cases when the inharmonious region is separated by background.}
     \label{limitation}
 \end{figure}

We have found that for a few cases when the inharmonious region is separated into several parts, our method may fail to detect some parts which are very small (see Figure~\ref{limitation}) and only detect the part which appears to be the most inharmonious. In such cases, context information may need to be considered to segment the region completely, which is left for future work.


